\documentclass[11pt,a4paper]{article}
\usepackage[hyperref]{acl2020}
\usepackage{times}
\usepackage{latexsym}

\usepackage{xspace}
\usepackage{url}
\usepackage{multirow}
\usepackage{booktabs}
\usepackage{graphicx}
\usepackage{amsmath}
\usepackage{bbm}
\usepackage{amssymb}
\usepackage{arydshln}

\usepackage{microtype}

\aclfinalcopy 
\def\aclpaperid{2435} 

\usepackage{xcolor}
\usepackage[normalem]{ulem} 
\newcommand\hl{\bgroup\markoverwith
  {\textcolor[RGB]{255,252,121}{\rule[-.5ex]{2pt}{2.5ex}}}\ULon}
\newcommand\tl{\bgroup\markoverwith
  {\textcolor[RGB]{255,126,121}{\rule[-.5ex]{2pt}{2.5ex}}}\ULon}
\newcommand\rl{\bgroup\markoverwith
  {\textcolor[RGB]{115,250,121}{\rule[-.5ex]{2pt}{2.5ex}}}\ULon}
\newcommand\gl{\bgroup\markoverwith
  {\textcolor[rgb]{0.8,0.8,0.8}{\rule[-.5ex]{2pt}{2.5ex}}}\ULon}
\newcommand\MODELNAME{RecoverSAT}
\newcommand\SemiNAT{\MODELNAME\xspace}

\title{Learning to Recover from Multi-Modality Errors for Non-Autoregressive\\Neural Machine Translation}

\author{Qiu Ran\thanks{\ \ indicates equal contribution}, Yankai Lin\footnotemark[1]$\,$  \thanks{\ \ indicates corresponding author}, Peng Li\footnotemark[1]$\,$  \footnotemark[2], Jie Zhou \\
  Pattern Recognition Center, WeChat AI, Tencent Inc., China \\
  \texttt{\{soulcaptran,yankailin,patrickpli,withtomzhou\}@tencent.com} \\}

\date{}

\begin{document}

\maketitle
\begin{abstract}
Non-autoregressive neural machine translation (NAT) predicts the entire target sequence simultaneously and significantly accelerates inference process. However, NAT discards the dependency information in a sentence, and thus inevitably suffers from the multi-modality problem: the target tokens may be provided by different possible translations, often causing token repetitions or missing. To alleviate this problem, we propose a novel semi-autoregressive model \SemiNAT in this work, which generates a translation as a sequence of segments. The segments are generated simultaneously while each segment is predicted token-by-token. By dynamically determining segment length and deleting repetitive segments, \SemiNAT is capable of recovering from repetitive and missing token errors. Experimental results on three widely-used benchmark datasets show that our proposed model achieves more than 4$\times$ speedup while maintaining comparable performance compared with the corresponding autoregressive model.
\end{abstract}

\section{Introduction}

Although neural machine translation (NMT) has achieved state-of-the-art performance in recent years~\citep{cho2014learning,bahdanau2015neural,vaswani2017attention}, most NMT models still suffer from the slow decoding speed problem due to their autoregressive property: the generation of a target token depends on all the previously generated target tokens, making the decoding process intrinsically nonparallelizable.

Recently, non-autoregressive neural machine translation (NAT) models~\citep{gu2018non,li2019hint,wang2019non,guo2019non,wei2019imitation} have been investigated to mitigate the slow decoding speed problem by generating all target tokens independently in parallel, speeding up the decoding process significantly. Unfortunately, these models suffer from the multi-modality problem~\citep{gu2018non}, resulting in inferior translation quality compared with autoregressive NMT. To be specific, a source sentence may have multiple feasible translations, and each target token may be generated with respect to different feasible translations since NAT models discard the dependency among target tokens. This generally manifests as repetitive or missing tokens in the translations. Table~\ref{tab:multi-modality} shows an example. The German phrase ``{\it viele Farmer}'' can be translated as either ``\hl{{\it lots of farmers}}'' or ``\rl{{\it a lot of farmers}}''. In the first translation (Trans. 1), ``\hl{{\it lots of}}'' are translated w.r.t. ``\hl{{\it lots of farmers}}'' while ``\rl{{\it of farmers}}'' are translated w.r.t. ``\rl{{\it a lot of farmers}}'' such that two ``{\it of}'' are generated. Similarly, ``{\it of}'' is missing in the second translation (Trans. 2). Intuitively, the multi-modality problem has a significant negative effect on the translation quality of NAT.

\begin{table}[!t]
  \small
  \centering
  \scalebox{0.9}{
  \begin{tabular}{lp{0.36\textwidth}}
    \toprule
    {\bf Src.} & es gibt heute \gl{viele Farmer} mit diesem Ansatz\\
    \midrule
    {\bf Feasible} & there are \hl{lots of farmers} doing this today\\
    {\bf  Trans.} & there are \rl{a lot of farmers} doing this today\\
\midrule
    {\bf Trans. 1} &   there are \hl{lots of} \rl{of farmers} doing this today\\
    {\bf Trans. 2} &   there are \rl{a lot} \hl{farmers} doing this today\\
    \bottomrule
    \end{tabular}}
    \caption{A multi-modality problem example: NAT models generate each target token independently such that they may correspond to different feasible translations, which usually manifests as repetitive (Trans. 1) or missing (Trans. 2) tokens.
    }
    \label{tab:multi-modality}
\end{table}

\begin{figure*}
    \centering
    \includegraphics[width=0.9\textwidth]{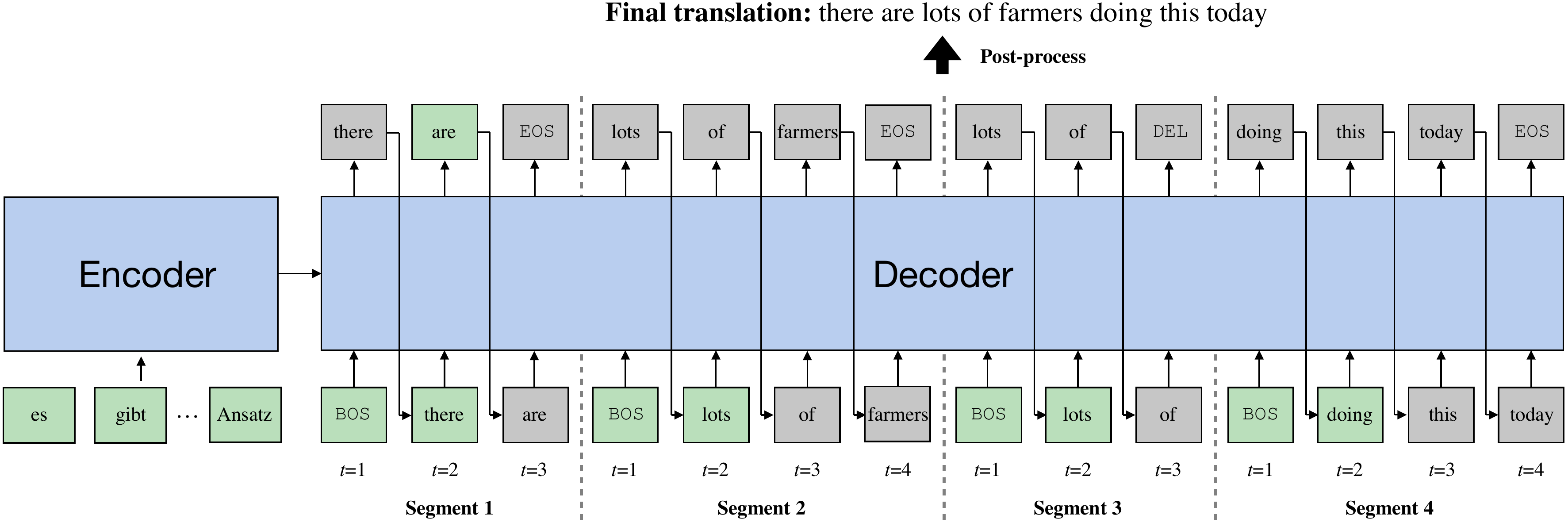}
    \caption{An overview of our \SemiNAT model. \SemiNAT generates a translation as a sequence of segments. The segments are generated simultaneously while each segment is generated token-by-token conditioned on both the source tokens and the translation history of all segments (e.g., the token ``{\em are}'' in the first segment is predicted based on all the tokens colored green). Repetitive segments (e.g., the third segment ``{\em lots of}'') are detected and deleted automatically.}
    \label{fig:model} 
\end{figure*}

Intensive efforts have been devoted to alleviate the above problem, which can be roughly divided into two lines.
The first line of work leverages the iterative decoding framework to break the independence assumption, which first generates an initial translation and then refines the translation iteratively by taking both the source sentence and the translation of last iteration as input~\citep{lee2018deterministic,ghazvininejad2019mask}. Nevertheless, it requires to refine the translations for multiple times in order to achieve better translation quality, which hurts decoding speed significantly.
The other line of work tries to improve the vanilla NAT model to better capture target-side dependency by leveraging extra autoregressive layers in the decoder~\cite{shao2019retrieving,wang2018semi}, introducing latent variables and/or more powerful probabilistic frameworks to model more complex distributions~\cite{kaiser2018fast,akoury2019syntactically,shu2019latent,ma2019flowseq}, guiding the training process with an autoregressive model~\cite{li2019hint,wei2019imitation}, etc.
However, these models cannot alter a target token once it has been generated, which means these models are not able to recover from an error caused by the multi-modality problem.

To alleviate the multi-modality problem while maintaining a reasonable decoding speedup, we propose a novel semi-autoregressive model named \SemiNAT in this work. 
\SemiNAT features in three aspects: (1) To improve decoding speed, we assume that a translation can be divided into several segments which can be generated simultaneously. (2) To better capture target-side dependency, the tokens inside a segment is autoregressively generated conditioned not only on the previously generated tokens in this segment but also on those in other segments. On one hand, we observe that repetitive tokens are more likely to occur within a short context. Therefore, autoregressively generating a segment is beneficial for reducing repetitive tokens. On the other hand, by conditioning on previously generated tokens in other segments, the model is capable of guessing what feasible translation candidates have been chosen by each segment and adapts accordingly, e.g., recovering from missing token errors. As a result, our model captures more target-side dependency such that the multi-modality problem can be alleviated naturally.
(3) To make the model capable of recovering from repetitive token errors, we introduce a segment deletion mechanism into our model. Informally speaking, our model will mark a segment to be deleted once it finds the content has been translated in other segments.

We conduct experiments on three benchmark datasets for machine translation to evaluate the proposed method. The experimental results show that \SemiNAT is able to decode over 4$\times$ faster than the autoregressive counterpart while maintaining comparable performance. The source code of this work is released on \url{https://github.com/ranqiu92/\MODELNAME}.

\section{Background}

\subsection{Autoregressive Neural Machine Translation}

Autoregressive neural machine translation (AT) generates the translation token-by-token conditioned on translation history. Denoting a source sentence as $\mathbf{x}=\{x_i\}_{i=1}^{T'}$ and a target sentence as $\mathbf{y}=\{y_j\}_{j=1}^{T}$, AT models the joint probability as:
\begin{equation}
    P(\mathbf{y}|\mathbf{x})=\prod_{t=1}^{T}P(y_{t}|\mathbf{y}_{<t}, \mathbf{x}).
\end{equation}
where $\mathbf{y}_{<t}$ denotes the generated tokens before $y_t$.

During decoding, the translation history dependency makes the AT model predict each token after all previous tokens have been generated, which makes the decoding process time-consuming.

\subsection{Non-Autoregressive Neural Machine Translation}

Non-autoregressive neural machine translation (NAT)~\citep{gu2018non} aims to accelerate the decoding process, which discards the dependency of translation history and models $P(\mathbf{y}|\mathbf{x})$ as a product of the conditionally independent probability of each token:
\begin{equation}
    P(\mathbf{y}|\mathbf{x})=\prod_{t=1}^{T}P(y_{t}|\mathbf{x}).
\end{equation}
The conditional independence enables the NAT models to generate all target tokens in parallel.

However, independently predicting all target tokens is challenging as natural language often exhibits strong correlation across context. Since the model knows little information about surrounding target tokens, it may consider different possible translations when predicting different target tokens. The problem is known as the {\em multi-modality problem}~\citep{gu2018non} and significantly degrades the performance of NAT models.

\section{Approach}

\subsection{Overview}
\label{sec:model:overview}
\SemiNAT extends the original Transformer~\citep{vaswani2017attention} to enable the decoder to perform generation autoregressively in local and non-autoregressively in global. An overview of the architecture of our \SemiNAT model  is shown in Figure~\ref{fig:model}. As illustrated in the figure, \SemiNAT simultaneously predicts all segments ``{\em there are} \texttt{EOS}'', ``{\em lots of farmers} \texttt{EOS}'', ``{\em a lot} \texttt{DEL}'' and ``{\em doing this today} \texttt{EOS}''. And at each time step, it generates a token for each incomplete segment. The special token \texttt{DEL} denotes the segment should be deleted and \texttt{EOS} denotes the end of a segment. Combining all the segments, we obtain the final translation ``{\em there are lots of farmers doing this today}''.

Formally, assuming a translation $\mathbf{y}$ is generated as $K$ segments $\mathbf{S}^1,\mathbf{S}^2,\cdots,\mathbf{S}^{K}$, where $\mathbf{S}^i$ is a sub-sequence of the translation\footnote{Note that, by fixing segment length (token number of each segment) instead, the segment number $K$ can be changed dynamically according to the sentence length. In other words, we can predict the target sentence length to determine the segment number during inference. In this case, our model can also decode in constant time.}. For description simplicity, we assume that all the segments have the same length.
\SemiNAT predicts a token for each segment conditioned on all previously generated tokens at each generation step, which can be formulated as:
\begin{equation}
    P(\mathbf{y}|\mathbf{x})=\prod_{t=1}^L\prod_{i=1}^K P(\mathbf{S}^i_t|\mathbf{S}^1_{<t} \cdots \mathbf{S}^K_{<t};\mathbf{x}),
    \label{eqn:model}
\end{equation}
where $\mathbf{S}^i_t$ denotes the $t$-th token in the $i$-th segment, $\mathbf{S}^i_{<t}=\{\mathbf{S}^i_1,\cdots,\mathbf{S}^i_{t-1}\}$ denotes the translation history in the $i$-th segment, and $L$ is segment length.

Here, two natural problems arise for the decoding process: 
\begin{itemize}
    \item How to determine the length of a segment?
    \item How to decide a segment should be deleted?
\end{itemize}
We address the two problems in a uniform way in this work. Suppose the original token vocabulary is $V$, we extend it with two extra tokens \texttt{EOS} and \texttt{DEL}. Then for the segment $\mathbf{S}^i$, the most probable token $\hat{\mathbf{S}}^i_t$ at time step $t$:
\begin{equation}
    \hat{\mathbf{S}}^i_t=\mathop{\arg\max}_{\mathbf{S}^i_t\in V\cup\{\texttt{EOS},\texttt{DEL}\}}P(\mathbf{S}^i_t|\mathbf{S}^1_{<t} \cdots \mathbf{S}^K_{<t};\mathbf{x})
\end{equation}
has three possibilities:

(1)  $\hat{\mathbf{S}}^i_t\in V$: the segment $\mathbf{S}^i$ is incomplete and the decoding process for it should continue;

(2) $\hat{\mathbf{S}}^i_t=\texttt{EOS}$: the segment $\mathbf{S}^i$ is complete and the decoding process for it should terminate;

(3)  $\hat{\mathbf{S}}^i_t=\texttt{DEL}$: the segment $\mathbf{S}^i$ is repetitive and should be deleted. Accordingly, the decoding process for it should terminate.

The entire decoding process terminates when all the segments meet \texttt{EOS}/\texttt{DEL} or reach the maximum token number.
It should be noticed that we do not explicitly delete a segment when \texttt{DEL} is encountered but do it via post-processing. In other words, the model is trained to ignore the segment to be deleted implicitly.

\subsection{Learning to Recover from Errors}
As there is little target-side information available in the early stage of the decoding process, the errors caused by the multi-modality problem is inevitable. In this work, instead of reducing such errors directly, we propose two training mechanisms to teach our \SemiNAT model to recover from errors: (1) Dynamic Termination Mechanism: learning to determine segment length according to target-side context; (2) Segment Deletion Mechanism: learning to delete repetitive segments.

\subsubsection{Dynamic Termination Mechanism}
\label{sec:dynamic-termination}
As shown in Section~\ref{sec:model:overview}, instead of pre-specifying the lengths of segments, we let the model determine the lengths by emitting the \texttt{EOS} token. This strategy helps our model recover from multi-modality related errors in two ways:

1.  The choice of the first few tokens is more flexible. Taking Figure~\ref{fig:model} as an example, if the decoder decides the first token of the second segment is ``{\em of}'' instead of ``{\em lots}'' (i.e., ``{\em lots}'' is not generated in the second segment), it only needs to generate ``{\em lots}'' before ``\texttt{EOS}'' in the first segment in order to recover from missing token errors. In contrast, if the decoder decides the first token is ``{\em are}'', it can avoid repetitive token error by not generating ``{\em are}'' in the first segment;

2.  As shown in Eq.~\ref{eqn:model}, a token is generated conditioned on all the previously generated tokens in {\em all the segments}. Therefore, the decoder has richer target-side information to detect and recover from such errors.

However, it is non-trivial to train the model to learn such behaviour while maintaining a reasonable speedup. On one hand, as the decoding time of our \SemiNAT model is proportional to the maximum length of the segments, we should divide the target sentences of training instances into equal-length segments to encourage the model to generate segments with identical length. On the other hand, the model should be exposed to the multi-modality related errors to enhance its ability of recovering from such errors, which suggests that the target sentences of training instances should be divided randomly to simulate these errors.

To alleviate the problem, we propose a mixed annealing dividing strategy. To be specific, we randomly decide whether to divide a target sentence equally or randomly at each training step and gradually anneal to the equally-dividing method at the end of training. Formally, given the target sentence $\mathbf{y}$ and the segment number $K$, we define the segment dividing indice set $\mathbf{r}$ as follows:
\begin{eqnarray}
    s&\sim&\mathrm{Bernoulli}(p),\label{eqn:bernoulli}\\
    \mathbf{r}&=&
        \begin{cases}
              \mathrm{EQUAL}(T, K-1) & s = 0\\
              \mathrm{RAND}(T, K-1) & s = 1
        \end{cases},
\end{eqnarray}
where $\mathrm{Bernoulli}(p)$ is the Bernoulli distribution with parameter $p$, $\mathrm{EQUAL}(n,m)=\big\{\lceil{\frac{n}{m+1}}\rceil, \lceil{\frac{2n}{m+1}}\rceil,\cdots,\lceil{\frac{mn}{m+1}}\rceil\big\}$, $\mathrm{RAND}(n,m)$ sampling $m$ non-duplicate indices from $[1, n]$. 
A larger value of $p$ leads to better error recovering ability while a smaller one encourages the model to generate segments with similar lengths (in other words, better speedup). To balance the two aspects, we gradually anneal $p$ from $1$ to $0$ in the training process, which achieves better performance (Section~\ref{sec:termination}).
 
\subsubsection{Segment Deletion Mechanism}
\label{sec:segment-deletion}

Although the dynamic termination mechanism makes the model capable of recovering from missing token errors and reducing repetitive tokens, the model still can not recover from errors where token repetition errors have already occurred. We find the major errors of our model occur when generating the first token of each segment since it cannot see any history and future. In this situation, two repetitive segments will be generated. To alleviate this problem, we propose a segment-wise deletion strategy, which uses a special token \texttt{DEL} to indicate a segment is repetitive and should be deleted\footnote{It is more flexible to employ token-wise deletion strategy which could handle more complex cases. We will explore this in future.}. 

A straightforward way to train the model to learn to delete a segment is to inject pseudo repetitive segments into the training data. The following is an example:
\begin{table}[h]
  \small
  \centering
  \begin{tabular}{p{0.33\columnwidth} p{0.25\textwidth}}
    \toprule
    Target Sentence & there are lots of farmers doing this today\\
    \midrule
    + Pseudo Repetitive Segment & \tl{there are} \hl{lots of farmers} \gl{lots of \texttt{DEL}} \rl{doing this today}\\
    \bottomrule
    \end{tabular}
    \label{tab:deletion_policy_construction}
\end{table}

\noindent Given the target sentence ``{\em there are lots of farmers doing this today}'', we first divide it into $3$ segments ``{\em there are}'', ``{\em lots of farmers}'' and ``{\em doing this today}''. Then we copy the first two tokens of the second segment and append the special token \texttt{DEL} to the end to construct a pseudo repetitive segment ``{\em lots of} \texttt{DEL}''. Finally, we insert the repetitive segment to the right of the chosen segment, resulting in $4$ segments.
Formally, given the expected segment number $K$ and the target sentence $\mathbf{y}$, we first divide $\mathbf{y}$ into $K-1$ segments $\mathbf{S}^1,\mathbf{S}^2,\cdots,\mathbf{S}^{K-1}$ and then build a pseudo repetitive segment $\mathbf{S}^i_{rep}$ by copying the first $m$ tokens of a randomly chosen segment $\mathbf{S}^i$ and appending \texttt{DEL} to the end, $m$ is uniformly sampled from $[1, |\mathbf{S}^i|]$. Finally, $\mathbf{S}^i_{rep}$ is inserted at the right side of $\mathbf{S}^i$. The final $K$ segments are $\mathbf{S}^1,\mathbf{S}^2,\cdots,\mathbf{S}^i,\mathbf{S}^i_{rep},\mathbf{S}^{i+1},\cdots,\mathbf{S}^{K-1}$.

However, injecting such pseudo repetitive segments to all training instances will mislead the model that generating then deleting a repetitive segment is a must-to-have behaviour, which is not desired. Therefore, we inject pseudo repetitive segment into a training instance with probability $q$ in this work.

\begin{table*}[!t]
  \centering
  \small
  \resizebox{0.97\textwidth}{!}{
  \begin{tabular}{lcccr|ccr|cr}
    \toprule
    \multicolumn{1}{l}{\multirow{2}{*}{Model}}  & Iterative & \multicolumn{3}{c|}{WMT14 En-De}     & \multicolumn{3}{c|}{WMT16 En-Ro}     & \multicolumn{2}{c}{IWSLT16 En-De}\\
      & Decoding & En$\to$     & De$\to$  & Speedup & En$\to$     & Ro$\to$  & Speedup &  En$\to$     &     Speedup                  \\
    \midrule
    Transformer             &  & 27.17         & 31.95            & 1.00$\times$	        & 32.86         & 32.60             & 1.00$\times$         & 31.18         & 1.00$\times$\\
    \midrule
    NAT-FT+NPD ($n=100$)     &  & 19.17         & 23.20       & -         & 29.79         & 31.44        & -         & 28.16         & 2.36$\times$\\
    SynST                  &  & 20.74         & 25.50       & 4.86$\times$         &  -              & -           & -          & 23.82                 & 3.78$\times$\\
    NAT-IR ($iter=10$)     & \checkmark & 21.61         & 25.48     & 2.01$\times$        & 29.32         & 30.19     & 2.15$\times$         & 27.11         & 1.55$\times$\\
    NAT-FS                 &  & 22.27         & 27.25       & 3.75$\times$         & 30.57         & 30.83     &3.70$\times$         & 27.78         & 3.38$\times$\\
    imitate-NAT+LPD ($n=7$)       & & 24.15         & 27.28   & -	       & 31.45	       & 31.81    & -         & 30.68	       & 9.70$\times$\\
    PNAT+LPD ($n=9$)     & & 24.48         & 29.16         & -           & -            & -            & -    & -   & -   \\
    NAT-REG+LPD ($n=9$)  & & 24.61         & 28.90         & -           & -            & -   & -     & 27.02         & -\\
    LV NAR            & & 25.10         & -              & 6.8$\times$   & -            & -             & -             & -             & -\\
    NART+LPD ($n=9$)     & & 25.20         & 29.52     & 17.8$\times$     & -         & -          & -             & -             & -\\
    FlowSeq+NPD ($n=30$)  & & 25.31      & 30.68         & $<$1.5$\times$        & 32.20         & 32.84         & -   & -    & -\\
    FCL-NAT+NPD ($n=9$)  & & 25.75         & 29.50             & 16.0$\times$          & -             & -            & -    & -   & -\\
    ReorderNAT             & & 26.51         & 31.13         & -         & 31.70         & 31.99     & -         & 30.26        & 5.96$\times$\\
    NART-DCRF+LPD ($n=19$)   & & 26.80       & 30.04           & 4.39$\times$          & -             & -         & -        & -   & -\\
    SAT ($K=2$)              & & 26.90         & -           & 1.51$\times$          & -             & -        & -         & -    & -\\
    CMLM ($iter=10$)  & \checkmark & 27.03         & 30.53   & $<$1.5$\times$    & \textbf{33.08}  & \textbf{33.31}    & -     & -    & -\\
    \midrule
    \SemiNAT ($K=2$)    & & \textbf{27.11}   & \textbf{31.67}    & 2.16$\times$      & 32.92         &  33.19    & 2.02$\times$        & \textbf{30.78}   & 2.06$\times$\\
    \SemiNAT ($K=5$)    & & 26.91           & 31.22     & 3.17$\times$     & 32.81         &  32.80   & 3.16$\times$        & 30.55   & 3.28$\times$\\
    \SemiNAT ($K=10$)   &  & 26.32          & 30.46     & 4.31$\times$     &  32.59        & 32.29    & 4.31$\times$        & 29.90   & 4.68$\times$\\
    \bottomrule
    \end{tabular}}
    \caption{Performance (BLEU) of Transformer, the NAT/semi-autoregressive models and \SemiNAT on three widely-used machine translation benchmark datasets. NPD denotes the noisy parallel decoding technique~\citep{gu2018non} and LPD denotes the length parallel decoding technique~\citep{wei2019imitation}. $n$ denotes the sample size of NPD or LPD. $iter$ denotes the refinement number of the iterative decoding method.}
    \label{tab:main_results}
\end{table*}

\section{Experiments}

\subsection{Datasets}

We conduct experiments on three widely-used machine translation datasets: IWSLT16 En-De ($196$k pairs), WMT14 En-De ($4.5$M pairs) and WMT16 En-Ro ($610$k pairs). 
For fair comparison, we use the preprocessed datasets in~\citet{lee2018deterministic}, of which sentences are tokenized and segmented into subwords using byte-pair encoding (BPE)~\citep{sennrich2016neural} to restrict the vocabulary size. We use a shared vocabulary of $40$k subwords for both source and target languages. For the WMT14 En-De dataset, we use newstest-2013 and newstest-2014 as validation and test sets respectively. For the WMT16 En-Ro dataset, we employ newsdev-2016 and newstest-2016 as validation and test sets respectively. For the IWSLT16 En-De dataset, we use test2013 as the validation set.

\subsection{Experimental Settings}

For model hyperparameters, we follow most of the settings in~\citep{gu2018non,lee2018deterministic,wei2019imitation}. For the IWSLT16 En-De dataset, we use a small Transformer model ($d_{model}=278$, $d_{hidden}=507$, $n_{layer}=5$, $n_{head}=2$, $p_{dropout}=0.1$). For the WMT14 En-De and WMT16 En-Ro datasets, we use a larger Transformer model ($d_{model}=512$, $d_{hidden}=512$, $n_{layer}=6$,  $n_{head}=8$, $p_{dropout}=0.1$). We linearly anneal the learning rate from $3 \times 10^{-4}$ to $10^{-5}$ as in~\citet{lee2018deterministic} for the IWSLT16 En-De dataset, while employing the warm-up learning rate schedule~\citep{vaswani2017attention} with $t_{warmup}=4000$ for the WMT14 En-De and WMT16 En-Ro datasets. We also use label smoothing of value $\epsilon_{ls}=0.15$ for all datasets. We utilize the sequence-level distillation~\citep{kim2016sequence}, which replaces the target sentences in the training dataset with sentences generated by an autoregressive model, and set the beam size of the technique to $4$. We use the encoder of the corresponding autoregressive model to initialize the encoder of \SemiNAT, and share the parameters of source and target token embedding layers and the pre-softmax linear layer. We measure the speedup of model inference in each task on a single NVIDIA P40 GPU with the batch size $1$.

\subsection{Baselines}

We use the Transformer~\citep{vaswani2017attention} as our AT baseline and fifteen latest strong NAT models as NAT baselines, including: 
(1) fertility-based model: NAT-FT~\citep{gu2018non}; 
(2) iterative decoding based models: NAT-IR~\citep{lee2018deterministic} and CMLM~\citep{ghazvininejad2019mask}; 
(3) models learning from AT teachers: imitate-NAT~\citep{wei2019imitation}, NART~\citep{li2019hint} and FCL-NAT~\cite{guo2019fine}; 
(4) latent variable framework based models: LV NAR~\citep{shu2019latent} and FlowSeq~\citep{ma2019flowseq}; 
(5) regularization framework based model: NAT-REG~\citep{wang2019non};
(6) models introducing extra target-side dependencies: SAT~\citep{wang2018semi}, SynST~\citep{akoury2019syntactically}, NAT-FS~\citep{shao2019retrieving}, PNAT~\citep{bao2019non}, NART-DCRF~\citep{sun2019fast} and ReorderNAT~\citep{ran2019guiding}.

\subsection{Overall Results}

The performance of our \SemiNAT model and the baselines is shown in Table~\ref{tab:main_results}. Due to the space limitation, we only show the results corresponding to the settings of the best BLEU scores for the baselines \footnote{A thorough comparison under other settings can be found in Appendix B.}.
From Table~\ref{tab:main_results}, we can observe that:

(1) Our \SemiNAT model achieves comparable performance with the AT baseline (Transformer) while keeping significant speedup. When $K=2$, the BLEU score gap is moderate (from $0.06$ to $0.4$, even better than Transformer on the WMT16 En$\to$Ro and Ro$\to$En tasks) and the speedup is about $2\times$. When $K=10$, the BLEU scores drop less than $5\%$ relatively, and the speedup is considerably good (over $4\times$). 

(2) Our \SemiNAT model outperforms all the strong NAT baselines except CMLM (on the WMT16 En$\to$Ro and Ro$\to$En tasks). However, the performance gap is negligible ($0.16$ and $0.12$ respectively), and CMLM is a multi-step NAT method which is significantly slower than our model.

(3) As $K$ grows, the BLEU scores drop moderately and the speedup grows significantly, indicating that our \SemiNAT model has a good generalizability. For example, the BLEU scores drop less than $0.45$ when $K$ grows from $2$ to $5$, and drop no more than $0.90$ except on the WMT14 De$\to$En task when $K$ further grows to $10$. Meanwhile, the speedup for $K=10$ is larger than $4\times$, which is considerably good.

(4) There are only $7$ baselines (SynST, imitate-NAT+LPD, LV NAR, NART+LPD, FCL-NAT+NPD, ReorderNAT and NART-DCRF+LPD) achieving better speedup than our \SemiNAT model when $K=10$. However, only ReorderNAT and NART-DCRF+LPD achieve comparable BLEU scores with our model.
The improvements of both ReorderNAT and NART-DCRF are complementary to our method. It is an interesting future work to join these works together.

\subsection{Effect of Dynamic Termination Mechanism}
\label{sec:termination}
As discussed in Section~\ref{sec:dynamic-termination}, the dynamic termination mechanism is used to train our \SemiNAT model to learn to determine segment length dynamically conditioned on target-side context such that it is recoverable from multi-modality related errors. In this section, we investigate the effect of this mechanism and the results are shown in Table~\ref{tab:segment_splitting}.

As multi-modality related errors generally manifest as repetitive or missing tokens in the translation, we propose two quantitative metrics ``Rep'' and ``Mis'' to measure these two phenomenons respectively. ``Rep'' is defined as the relative increment of repetitive token ratio w.r.t. to a reference AT model. And ``Mis'' is defined as the relative increment of missing token ratio given the references w.r.t. to a reference AT model. Formally, given the translations  $\hat{\mathbf{Y}}=\{\hat{\mathbf{y}}^1\cdots\hat{\mathbf{y}}^k\cdots\}$ produced by the model to be evaluated and the translations $\hat{\mathbf{Y}}_{auto}=\{\hat{\mathbf{y}}_{auto}^1\cdots\hat{\mathbf{y}}_{auto}^k\cdots\}$ produced by the reference AT model, ``Rep'' is defined as
\begin{equation}
       \text{Rep}=\frac{r(\hat{\mathbf{Y}}) - r(\hat{\mathbf{Y}}_{auto})}{r(\hat{\mathbf{Y}}_{auto})},
\end{equation}
\vspace{-1em}
\begin{equation}
   r(\mathbf{Y})=\frac{\sum\limits_k\sum\limits_{j=2}^{|\mathbf{y}^k|}\mathbbm{1}\left(\sum\limits_{i=1}^9\mathbbm{1}(\mathbf{y}^k_j=\mathbf{y}^k_{j-i})\ge 1\right)}{\sum\limits_k{|\mathbf{y}^k|}},
\end{equation}
where $\mathbbm{1}(cond)=1$ if the condition $cond$ holds otherwise $0$, and $\mathbf{y}^k_j$ is the $j$-th token of the translation sentence $\mathbf{y}^k$.

Given $\hat{\mathbf{Y}}$,  $\hat{\mathbf{Y}}_{auto}$ and references $\bar{\mathbf{Y}}=\{\bar{\mathbf{y}}^1\cdots\bar{\mathbf{y}}^k\cdots\}$, ``Mis'' is defined as 
\begin{equation}
    \text{Mis}=\frac{m(\hat{\mathbf{Y}}, \bar{\mathbf{Y}}) -  m(\hat{\mathbf{Y}}_{auto}, \bar{\mathbf{Y}})}{m(\hat{\mathbf{Y}}_{auto}, \bar{\mathbf{Y}})},
\end{equation}
where $m(\cdot, \cdot)$ computes the missing token ratio and is defined as follows:
\begin{eqnarray}
    c_w(\mathbf{y}^k,\bar{\mathbf{y}}^k)&=&\max\left(c(\bar{\mathbf{y}}^k,w)
    -c(\mathbf{y}^k,w), 0\right), \nonumber\\
    m(\mathbf{Y}, \bar{\mathbf{Y}})&=&\frac{\sum_k\sum_{w \in \bar{\mathbf{y}}^k} c_w(\mathbf{y}^k,\bar{\mathbf{y}}^k)}{\sum_k|\bar{\mathbf{y}}^k|},
\end{eqnarray}
where $c(\mathbf{y}, w)$ is the occurrence number of a token $w$ in the sentence $\mathbf{y}$.

From Table~\ref{tab:segment_splitting}, we can observe that: (1) By using the dynamic termination mechanism ($p= 0.5$, $1.0$, $1\to0$, where $p$ is the parameter of Bernoulli distribution (Eq.~\ref{eqn:bernoulli})), both repetitive and missing token errors are reduced (``Rep'' \& ``Mis''), and the BLEU scores are increased, indicating the effectiveness of the mechanism; (2) As $p$ grows larger, the average number of decoding steps (``Step'') increases significantly.
The reason is that more target sentences are divided into segments equally with smaller $p$ during training and the model is biased to generate segments with similar lengths. However, if the model is not exposed to randomly divided segments ($p=0.0$), it fails to learn to recover from multi-modality related errors and the BLEU score drops significantly.
(3) By using the {\em annealing dividing strategy} ($p=1\to0$, see Section~\ref{sec:dynamic-termination}), we achieve a good balance between decoding speed and translation quality. Therefore, we use it as the default setting in this paper.
\begin{table}[!t]
  \centering
  \small
  \scalebox{1.}{
  \begin{tabular}{lcrrrr}
    \toprule
     & $p$   & BLEU   &  Rep  &   Mis  &   Step\\
    \midrule
    NAT &   &   24.57   &   50.09   &   9.09   &   1\\
    \midrule
        &   0.0   &   27.09   &   22.05   &   6.95   & 4.2\\
    \SemiNAT   &   0.5   &   29.80   &   12.69   &   3.96   &   5.5\\
    ($K$=10)   &    1.0    &   29.89   &   13.00   &   4.75   &   7.2\\
       & 1$\to$0  &   29.90   &   7.09   &   3.56   &   5.1\\
    \bottomrule
    \end{tabular}}
    \caption{Effect of the dynamic termination mechanism. The results are evaluated on the IWSLT16 En-De validation set. $p$ is the parameter of Bernoulli distribution in Eq.~\ref{eqn:bernoulli}. ``Rep'' and ``Mis'' measure the relative increment (\%) of repetitive and missing token ratios (see Section~\ref{sec:termination}), the smaller the better. ``Step'' denotes the average number of decoding steps. And ``1$\to$0'' denotes annealing $p$ from $1$ to $0$ linearly.}
    \label{tab:segment_splitting}
\end{table}

\begin{table}[!t]
  \centering
  \small
  \scalebox{1.}{ 
  \begin{tabular}{lcrrr}
    \toprule
     & $q$   & BLEU  & Rep &   Step\\
    \midrule
    NAT &         &   24.57   & 50.09 &   1\\
    \midrule
                &    0.0    &   28.56   &   26.24   & 4.4\\
                &   0.1   &   29.73   &   5.11   & 4.7\\
    \SemiNAT    &   0.3   &   29.61   &   7.71   & 5.1\\
    ($K=10$)    &   0.5   &   29.90   &   7.09   & 5.1\\
                &   0.7   &   29.76   &   11.47  & 5.2\\
                &   0.9   &   29.25   &   21.38  & 5.3\\
                &   1.0   &   29.13   &   20.55  & 5.2\\
    \bottomrule
    \end{tabular}}
    \caption{Effect of segment deletion mechanism. The results are evaluated on the IWSLT16 En-De validation set. $q$ is the probability of injecting pseudo repetitive segments to each training instance (see Section~\ref{sec:segment-deletion}).
    }
    \label{tab:segment_deletion}
\end{table}

\subsection{Effect of Segment Deletion Mechanism}

In this section, we investigate the effect of the segment deletion mechanism and the results are shown in Table~\ref{tab:segment_deletion}, where $q$ is the probability of injecting pseudo repetitive segments to each training instance. From the results we can observe that: (1) Without using the segment deletion mechanism ($q=0$), the BLEU score drops significantly and the repetitive token errors (``Rep'') increase drastically, indicating that the mechanism is effective for recovering from repetitive token errors. (2) As $q$ grows larger, the average number of decoding steps (``Step'') increases steadily because the model is misled that to generate then delete a repetitive segment is expected. Thus, $q$ should not be too large. (3) The repetitive token errors (``Rep'') increase drastically when $q>0.7$. We believe that the reason is that the pseudo repetitive segments are constructed randomly, making it hard to learn the underlying mapping.
(4) The model achieves the best performance with $q=0.5$. Therefore, we set $q=0.5$ in our experiments.

\begin{table*}[!t]
  \small
  \centering
  \resizebox{0.95\textwidth}{!}{
  \begin{tabular}{p{0.1\textwidth}p{0.09\textwidth}p{0.72\textwidth}}
    \toprule
    Source & & die er\_greif\_endste Abteilung ist das Denk\_mal f{\"u}r die Kinder , das zum Ged\_enken an die 1,5 Millionen Kinder , die in den Konzent\_rations\_lagern und Gas\_k\_ammern vernichtet wurden , erbaut wurde .\\
    \midrule
    Reference & & the most tragic section is the children's mem\_orial , built in memory of 1.5 million children killed in concentration camps and gas cham\_bers .\\
    \toprule
    NAT & Translation & the most tangible department \hl{department} \tl{\ \ } the monument \hl{monument} \tl{\ \ } the children , which was built \tl{\ \ } \hl{commem\_}commem\_orate 1.5 \hl{1.5} million children \tl{\ \ } were destroyed in the concentration camps and gas cham\_bers .\\
    \midrule
    \SemiNAT ($K=10$) & Translation & $\!${\bf \color{red}{A}}: $_{[1]}$the \texttt{EOS} $_{[2]}$most tangible department is the \texttt{EOS} $_{[3]}$monument for children \texttt{EOS} $_{[4]}$built to \texttt{EOS} $_{[5]}$commem\_orate the 1.5 \texttt{EOS} $_{[6]}$million children destroyed \texttt{EOS} $_{[7]}$in the concentration camps and \texttt{EOS} $_{[8]}$\gl{in \texttt{DEL}} $_{[9]}$gas \texttt{EOS} $_{[10]}$cham\_bers . \texttt{EOS}\\
    \cmidrule{2-3}
     & Forced Translation & $\!${\bf \color{red}{B}}: $_{[1]}$the \texttt{EOS} $_{[2]}$most tangible department is the \texttt{EOS} $_{[3]}$monument for children \texttt{EOS} $_{[4]}$built to \texttt{EOS} $_{[5]}$commem\_orate \texttt{EOS} $_{[6]}$\rl{the} 1.5 million children destroyed \texttt{EOS} $_{[7]}$in the concentration camps and \texttt{EOS} $_{[8]}$\gl{in \texttt{DEL}} $_{[9]}$gas \texttt{EOS} $_{[10]}$cham\_bers . \texttt{EOS}\\
    \cmidrule{3-3}
     &  & $\!${\bf \color{red}{C}}: $_{[1]}$the \texttt{EOS} $_{[2]}$most tangible department is the \texttt{EOS} $_{[3]}$monument for children \texttt{EOS} $_{[4]}$built to \texttt{EOS} $_{[5]}$commem\_orate the 1.5 million children \texttt{EOS} $_{[6]}$\rl{destroyed} \texttt{EOS} $_{[7]}$in concentration camps and \texttt{EOS} $_{[8]}$\gl{in \texttt{DEL}} $_{[9]}$gas \texttt{EOS} $_{[10]}$cham\_bers . \texttt{EOS}\\
     \cmidrule{3-3}
     & & $\!${\bf \color{red}{D}}: $_{[1]}$the \texttt{EOS} $_{[2]}$most tangible department is the \texttt{EOS} $_{[3]}$monument for children \texttt{EOS} $_{[4]}$built to \texttt{EOS} $_{[5]}$commem\_orate the 1.5 million children destroyed \texttt{EOS} $_{[6]}$\rl{in} the concentration camps and \texttt{EOS} $_{[7]}$\gl{in the \texttt{DEL}} $_{[8]}$\gl{in \texttt{DEL}} $_{[9]}$gas \texttt{EOS} $_{[10]}$cham\_bers . \texttt{EOS}\\
    \bottomrule
    \end{tabular}}
    \caption{Translation examples of NAT and \SemiNAT. ``Forced Translation'' denotes the generated sentence when we manually force the model to generate a certain token (colored \rl{green}) at a certain position. We use \hl{yellow} color to label repetitive tokens, \tl{red} color to label missing tokens, and \gl{gray} color to label the segments to be deleted. We use ``\_'' to concatenate sub-words and subscript numbers (e.g., [1]) to mark the beginning of each segment.}
    \label{tab:cases}
\end{table*}

\subsection{Performance over Sentence Lengths}

\begin{figure}[!t]
     \centering
     \includegraphics[width=0.8\columnwidth]{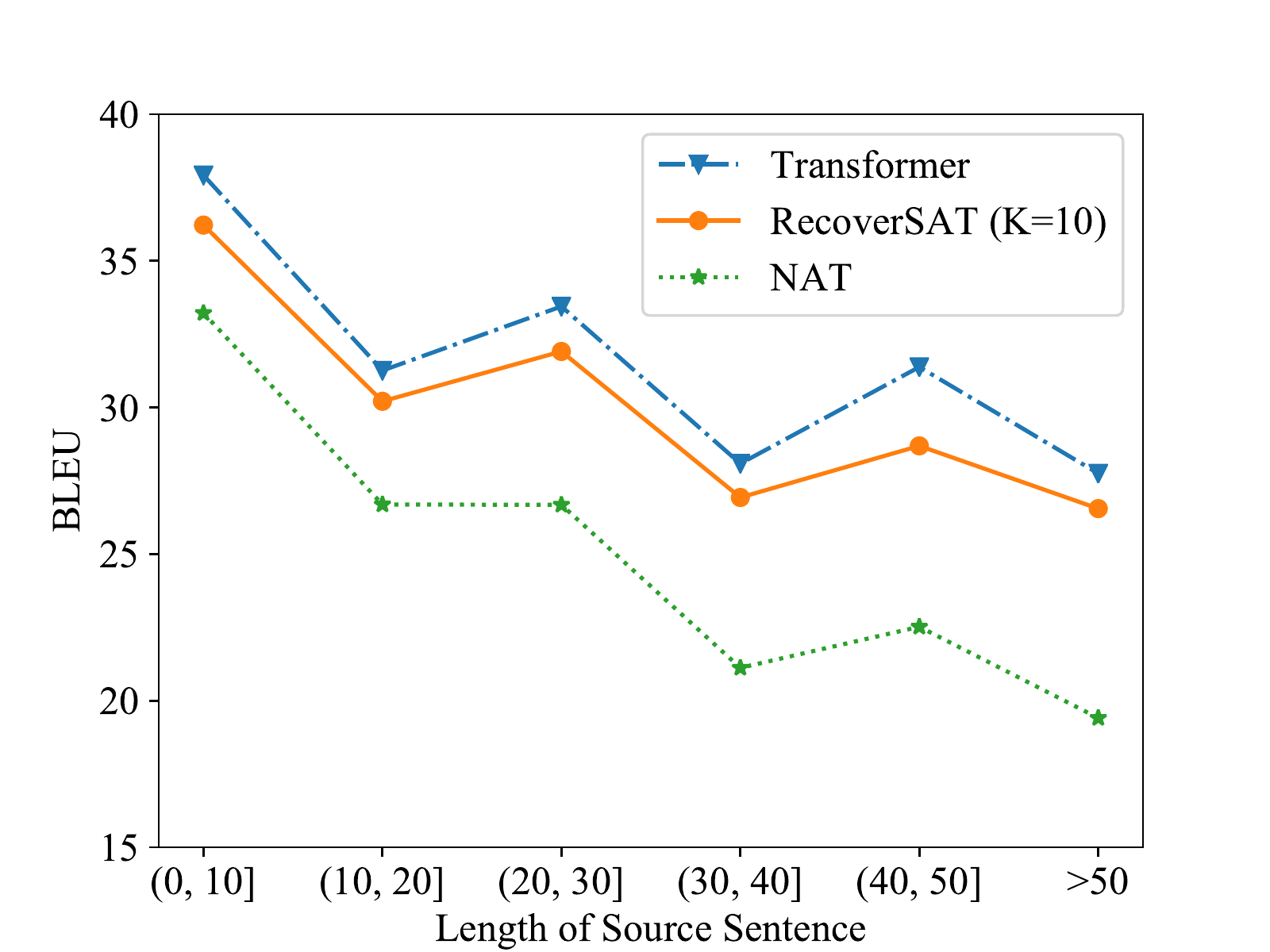}
     \caption{Translation quality on the IWSLT16 En-De validation set over sentences in different length.}
     \label{fig:bleu} 
\end{figure}

Figure~\ref{fig:bleu} shows the translation quality of the Transformer, our \SemiNAT model with $K=10$ and NAT on the IWSLT16 En-De validation set bucketed by different source sentence lengths. From the figure, we can observe that \SemiNAT surpasses NAT significantly and achieves comparable performance to the Transformer on all length buckets, which indicates the effectiveness of our model.

\subsection{Case Study}

We present translation examples of NAT and our \SemiNAT model on the WMT14 De$\to$En validation set in Table~\ref{tab:cases}. From the table, we can observe that: (1) The multi-modality problem (repetitive and missing tokens) is severe in the sentence generated by NAT, while it is effectively alleviated by \SemiNAT (see translations A to D); (2) \SemiNAT can leverage target contexts  to dynamically determine the segment length to reduce repetitive token errors (see translation B) or recover from missing token errors (see translations C and D); (3) \SemiNAT is capable of detecting and deleting the repetitive segments, even if there are multiple such segments (see translation D).

\section{Related Work}

There has been various work investigating to accelerate the decoding process of sequence generation models~\citep{kalchbrenner2018efficient,gu2018non}. In the field of neural machine translation, which is the focus of this work, \citet{gu2018non} first propose non-autoregressive machine translation (NAT), which generates all target tokens simultaneously.  Although accelerating the decoding process significantly, NAT suffers from the multi-modality problem~\citep{gu2018non} which generally  manifests as repetitive or missing tokens in translation. Therefore, intensive efforts have been devoted to alleviate the multi-modality problem in NAT.
\citet{wang2019non} regularize the decoder hidden states of neighboring tokens to reduce repetitive tokens;
\citet{sun2019fast} utilize conditional random field to model target-side positional contexts;
\citet{shao2019retrieving} and \citet{shao2019minimizing} introduce target-side information via specially designed training loss while \citet{guo2019non} enhance the input of the decoder with target-side information; \citet{kaiser2018fast}, \citet{akoury2019syntactically}, \citet{shu2019latent} and \citet{ma2019flowseq} incorporate latent variables to guide generation; \citet{li2019hint}, \citet{wei2019imitation} and \citet{guo2019fine} use  autoregressive models to guide the training process of NAT; 
\citet{ran2019guiding} and \citet{bao2019non} consider the reordering information in decoding.  \citet{wang2018semi} further propose a semi-autoregressive Transformer method, which generates segments autoregressively and predicts the tokens in a segment non-autoregressively. However, none of the above methods explicitly consider recovering from multi-modality related errors.

Recently, multi-step NAT models have also been investigated to address this issue. \citet{lee2018deterministic} and \citet{ghazvininejad2019mask} adopt an iterative decoding methods which have the potential to recover from generation errors. 
Besides, \citet{stern2019insertion} and \citet{gu2019levenshtein} also propose to use dynamic insertion/deletion to alleviate the generation repetition/missing. Different from these work, our model changes one-step NAT to a semi-autoregressive form, which maintains considerable speedup and enables the model to see the local history and future to avoid repetitive/missing words in decoding. Our work can further replace the one-step NAT to improve its performance.

\section{Conclusion}
In this work, we propose a novel semi-autoregressive model \SemiNAT to alleviate the multi-modality problem, which performs translation by generating segments non-autoregressively and predicts the tokens in a segment autoregressively. By determining segment length dynamically, \SemiNAT is capable of recovering from missing token errors and reducing repetitive token errors. By explicitly detecting and deleting repetitive segments, \SemiNAT is able to recover from repetitive token errors. Experiments on three widely-used benchmark datasets show that our \SemiNAT model maintains comparable performance with more than $4\times$ decoding speedup compared with the AT model.

\section*{Acknowledgments}
We would like to thank all anonymous reviewers for their insightful comments.

\bibliography{acl2020}
\bibliographystyle{acl_natbib}

\appendix

\begin{table*}[!t]
  \centering
  \small
  \resizebox{1.0\textwidth}{!}{
  \begin{tabular}{lcccr|ccr|cr}
    \toprule
    \multicolumn{1}{l}{\multirow{2}{*}{Model}}  & Iterative & \multicolumn{3}{c|}{WMT14 En-De}     & \multicolumn{3}{c|}{WMT16 En-Ro}     & \multicolumn{2}{c}{IWSLT16 En-De}\\
      & Decoding & En$\to$     & De$\to$  & Speedup & En$\to$     & Ro$\to$  & Speedup &  En$\to$     &     Speedup                  \\
    \midrule
    Transformer             &  & 27.17         & 31.95            & 1.00$\times$	        & 32.86         & 32.60             & 1.00$\times$         & 31.18         & 1.00$\times$\\
    \midrule
    NAT-FT                 &  & 17.69         & 21.47       & -         & 27.29         & 29.06        & -        & 26.52         & 15.6$\times$\\
    NAT-FT+NPD ($n=10$)      &  & 18.66         & 22.41       & -         & 29.02         & 30.76        & -        & 27.44         & 7.68$\times$\\
    NAT-FT+NPD ($n=100$)     &  & 19.17         & 23.20       & -         & 29.79         & 31.44        & -         & 28.16         & 2.36$\times$\\
    SynST                  &  & 20.74         & 25.50       & 4.86$\times$         &  -              & -           & -          & 23.82                 & 3.78$\times$\\
    NAT-IR ($iter=1$)      & \checkmark & 13.91         & 16.77     & 11.39$\times$        & 24.45         & 25.73     & 16.03$\times$         & 22.20         & 8.98$\times$\\
    NAT-IR ($iter=10$)     & \checkmark & 21.61         & 25.48     & 2.01$\times$        & 29.32         & 30.19     & 2.15$\times$         & 27.11         & 1.55$\times$\\
    NAT-FS                 &  & 22.27         & 27.25       & 3.75$\times$         & 30.57         & 30.83     &3.70$\times$         & 27.78         & 3.38$\times$\\
    imitate-NAT	           & & 22.44	       & 25.67      & -	       & 28.61	       & 28.90     & -         & 28.41	       & 18.6$\times$\\
    imitate-NAT+LPD ($n=7$)       & & 24.15         & 27.28   & -	       & 31.45	       & 31.81    & -         & 30.68	       & 9.70$\times$\\
    PNAT                   & & 23.05         & 27.18         & -             & -            & -            & -   & -   & -   \\
    PNAT+LPD ($n=9$)     & & 24.48         & 29.16         & -           & -            & -            & -    & -   & -   \\
    NAT-REG            & & 20.65         & 24.77         & -           & -            &-    &-      & 23.14         & -\\
    NAT-REG+LPD ($n=9$)  & & 24.61         & 28.90         & -           & -            & -   & -     & 27.02         & -\\
    LV NAR            & & 25.10         & -              & 6.8$\times$   & -            & -             & -             & -             & -\\
    NART                   & & 21.11         & 25.24         & 30.2$\times$         & -         & -             & -            & -            & -\\
    NART+LPD ($n=9$)     & & 25.20         & 29.52     & 17.8$\times$     & -         & -          & -             & -             & -\\
    FlowSeq-base           & & 21.45         & 26.16         & $<$1.5$\times$        & 29.34         & 30.44         & -   & -     & -\\
    FlowSeq-base+NPD ($n=30$)  & & 23.48       & 28.40         & $<$1.5$\times$        & 31.75         & 32.49         & -   & -     & -\\
    FlowSeq-large          & & 23.72         & 28.39         & $<$1.5$\times$        & 29.73         & 30.72         & -   & -     & -\\
    FlowSeq-large+NPD ($n=30$)  & & 25.31      & 30.68         & $<$1.5$\times$        & 32.20         & 32.84         & -   & -    & -\\
    FCL-NAT                & & 21.70         & 25.32         & 28.9$\times$          & -             & -            & -   & -    & -\\
    FCL-NAT+NPD ($n=9$)  & & 25.75         & 29.50             & 16.0$\times$          & -             & -            & -    & -   & -\\
    ReorderNAT             & & 26.51         & 31.13         & -         & 31.70         & 31.99     & -         & 30.26        & 5.96$\times$\\
    NART-DCRF            & & 23.44           & 27.22         & 10.4$\times$          & -             & -         & -        & -    & -\\
    NART-DCRF+LPD ($n=19$)   & & 26.80       & 30.04           & 4.39$\times$          & -             & -         & -        & -   & -\\
    SAT ($K=2$)              & & 26.90         & -           & 1.51$\times$          & -             & -        & -         & -    & -\\
    SAT ($K=6$)              & & 24.83         & -           & 2.98$\times$          & -             & -        & -         & -    & -\\
    CMLM-small ($iter=1$)  & \checkmark & 15.06         & 19.26   & -         & 20.12         & 20.36     & -         & -             & -\\
    CMLM-small ($iter=10$) & \checkmark & 25.51         & 29.47   & -         & 31.65         & 32.27     & -        & -             & -\\
    CMLM-base ($iter=1$)   & \checkmark & 18.05         & 21.83   & -         & 27.32         & 28.20     & -        & -             & -\\
    CMLM-base ($iter=10$)  & \checkmark & 27.03         & 30.53   & $<$1.5$\times$    & \textbf{33.08}  & \textbf{33.31}    & -     & -    & -\\
    \midrule
    \SemiNAT ($K=2$)    & & \textbf{27.11}   & \textbf{31.67}    & 2.16$\times$      & 32.92         &  33.19    & 2.02$\times$        & \textbf{30.78}   & 2.06$\times$\\
    \SemiNAT ($K=5$)    & & 26.91           & 31.22     & 3.17$\times$     & 32.81         &  32.80   & 3.16$\times$        & 30.55   & 3.28$\times$\\
    \SemiNAT ($K=10$)   &  & 26.32          & 30.46     & 4.31$\times$     &  32.59        & 32.29    & 4.31$\times$        & 29.90   & 4.68$\times$\\
    \bottomrule
    \end{tabular}}
    \caption{Performance (BLEU) of Transformer and the NAT/semi-autoregressive models on three widely-used machine translation benchmark datasets. NPD denotes the noisy parallel decoding technique~\citep{gu2018non} and LPD denotes the length parallel decoding technique~\citep{wei2019imitation}. $n$ denotes the sample size of NPD or LPD. $iter$ denotes the refinement number of the iterative decoding method.}
    \label{tab:main_results:more}
\end{table*}

\section{Positional Encoding}

Our \SemiNAT model utilizes the positional encoding method in \citet{vaswani2017attention} to encode the information about the positions of source tokens. The positional embedding is defined as:
\begin{eqnarray}
    \mathbf{PE}_{pos}[2i]&=&\sin\left(\frac{pos}{10000^{2i/d}}\right),\\
    \mathbf{PE}_{pos}[2i+1]&=&\cos\left(\frac{pos}{10000^{2i/d}}\right),
\end{eqnarray}
where $\mathbf{PE}_{pos}[i]$ is the $i$-th element of the positional embedding vector $\mathbf{PE}_{pos}$ for the position $pos$, and $d$ is the dimension of the positional embedding vector. Then we can compute the input vector of the encoder for the $m$-th source token $w$ as:
\begin{equation}
    \mathbf{E}_w= \mathbf{E}^{token}_w + \mathbf{PE}_m,
\end{equation}
where $\mathbf{E}^{token}_w$ is the token embedding vector of $w$.

However, we can not apply this method to {\em target} tokens directly. Since lengths of segments are dynamically determined, the positions of the tokens in the target sentence, except those in the first segment, are not available during generation. To solve the problem, we use the aforementioned method to independently encode the position in the corresponding segment of each token instead and adopt an absolute segment embedding method, which uses a distinct trainable vector to represent the position of each segment. Formally, the input vector of the decoder for the $n$-th target token $v$ of the $j$-th segment is computed as:
\begin{equation}
    \mathbf{E}_v = \mathbf{E}^{token}_v + \mathbf{PE}_n + \mathbf{E}^{seg}_j,
\end{equation}
where $\mathbf{E}^{seg}_j$ is the segment embedding vector for the segment position $j$.

\end{document}